\documentclass[journal]{IEEEtran}
\ifCLASSINFOpdf
   \usepackage[pdftex]{graphicx}
   \DeclareGraphicsExtensions{.pdf,.jpeg,.png}
\else
\fi
\usepackage{amsmath,amssymb}
\DeclareMathOperator*{\argmax}{arg\,max} % thin space, limits underneath in displays

\usepackage{subcaption}
\usepackage{multirow}
\usepackage[hidelinks]{hyperref}

\begin{document}
\title{Visual Diagnostics for Deep Reinforcement Learning Policy Development}

\newcommand{\printfnsymbol}[1]{%
  \textsuperscript{\@fnsymbol{#1}}%
}

\author{

\author{Jieliang~Luo\textsuperscript{*}\thanks{* Equal contribution.},
	Sam~Green\textsuperscript{*},
	Peter~Feghali,
	George~Legrady,
	and~\c{C}etin~Kaya~Ko\c{c}\\
	University of California, Santa Barbara\\
	jieliang@ucsb.edu}
    
}

\maketitle

\begin{abstract}
Modern vision-based reinforcement learning techniques often use convolutional neural networks (CNN) as universal function approximators to choose which action to take for a given visual input. Until recently, CNNs have been treated like black-box functions, but this mindset is especially dangerous when used for control in safety-critical settings. In this paper, we present our extensions of CNN visualization algorithms to the domain of vision-based reinforcement learning. We use a simulated drone environment as an example scenario. These visualization algorithms are an important tool for behavior introspection and provide insight into the qualities and flaws of trained policies when interacting with the physical world. A video may be seen at \href{https://sites.google.com/view/drlvisual}{https://sites.google.com/view/drlvisual}.%
\end{abstract}


% \fi
\IEEEpeerreviewmaketitle

\section{Introduction}
\IEEEPARstart{R}einforcement learning (RL) is a family of methods aimed at training an \textbf{agent} to collect rewards from an environment through trial-and-error approaches. Since the deep Q-network (DQN) algorithm was introduced in 2013, there has been a surge of interest in using convolutional neural networks (CNN) in vision-based RL algorithms \cite{mnih2013playing}. In the context of cyber-physical systems, vision-based RL has exciting potential to provide high levels of autonomy in applications like robotics, self-driving cars, and infrastructure inspection. However, CNNs are known to be opaque to debugging and RL's emphasis on trial-and-error demands rigorous behavioral verification before they may be allowed control over safety-critical physical systems. This work adapts CNN visualization techniques to the domain of RL.

Existing CNN visualization techniques attempt to visualize classes, provide decision attribution, or cluster inputs according to their resulting label. Techniques considered here include:
\begin{itemize}
    \item t-SNE maps \cite{maaten2008visualizing} -- Clusters similar inputs by the output classifications they trigger.
    \item Class visualization \cite{simonyan2013deep} -- Generates inputs which trigger specified classifications.
    \item Attribution visualization \cite{selvaraju2017grad} -- Identifies image regions most responsible for a classification decision.
\end{itemize}
These techniques are useful for understanding \textit{what} would cause a CNN to make a particular decision (through class visualization) and \textit{why} a CNN made a particular decision (through t-SNE and attribution visualization).

CNN visualizations have proven to be valuable for identifying strengths and weaknesses in a trained network. For example, the feature visualization for GoogLeNet's saxophone class indeed extracts a saxophone shaped object from the network. That is, the method generates an image, which, when input into the trained GoogLeNet CNN, will maximize the output probability of the saxophone class. However, when looking at the generated image, one can clearly see that the outline of a man has also been extracted from the network! The cause of this is the fact that CNNs, unlike humans, look at every pixel in the training set and will therefore extract all biases with which it was presented during training. 

Existing visualization methods are powerful tools for gaining understanding and trust during CNN image classification development, but they are not adequate by themselves for use with RL. For example CNN-based RL often uses a \textbf{stochastic policy} which means that an agent's action is chosen randomly under a distribution defined by the CNN's output (i.e. ``class'') probabilities. Visual diagnostics for stochastic policies should capture such uncertainty. Finally, existing CNN visualizations are designed for static images, not for the types of time-series data collected when an agent interacts with an environment.

\begin{figure}[!t]
\centering
\includegraphics[width=3.45in]{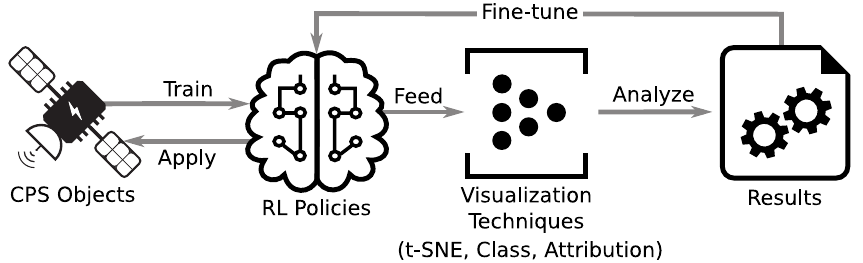}
\caption{Visualization development cycle when using convolutional neural networks for vision-based reinforcement learning. By iteratively visualizing \textit{what} and \textit{how} the CNN is perceiving, the engineer gains insight regarding \textit{why} the RL policy makes its decisions.}
\label{fig:cycle}
\end{figure}

We have adapted CNN visualization techniques to the domain of CNN-based RL. We show that these tools are valuable for providing explanations regarding an agent's decision-making process and can help an engineer understand policy deficiencies. In the following section, we provide: 1) a formal introduction to the goals of RL, our methodology, and related work; 2) visualization results for simulated drone experiments; 3) conclusions and opportunities for further research.

\section{Problem formulation and related work}

We have extended AirSim, a drone simulator created by Microsoft \cite{airsim2017fsr2}. AirSim is an Unreal Engine plugin which provides physically realistic simulations for autonomous vehicles. Our extensions add support for a variety of reinforcement learning tasks.

\begin{figure}[!t]
\centering
\includegraphics[width=3.45in]{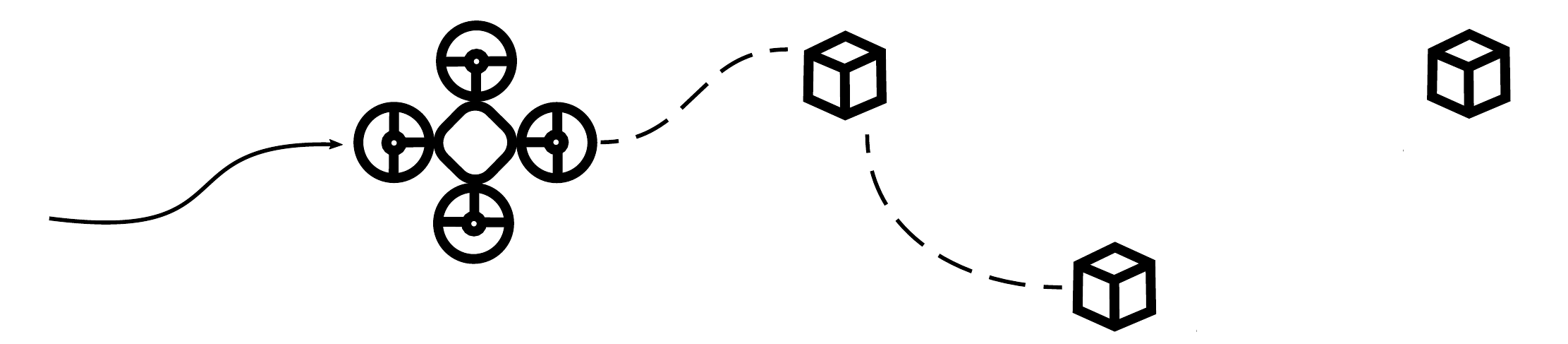}
\caption{Illustration of cube-collection environment. The drone policy is rewarded for ``collecting'' as many boxes as possible in each episode.}
\label{fig:airsim}
\end{figure}

As illustrated in Fig.~\ref{fig:airsim}, our AirSim environment features a drone which learns how to maneuver in order to ``collect'' cubes which are in front of it. At the beginning of each episode, the drone is reset to a starting point, and cubes are randomly distributed in front of it. The drone is controlled by a policy which is rewarded for collecting as many cubes as possible in each episode.

Formally, at each time step $t$, the drone's camera receives a partial observation of its state $s_t$ and then makes an action $a_t$. The agent's policy $\pi_\theta :s_t \rightarrow a_t$ is the logic with takes state observations and returns action selections. The possible actions in our environment are: ``left'', ``forward'', and ``right''. After each action the environment will return a new state observation $s_{t+1}$ and reward $r_{t+1}$. The policy is represented as a convolutional neural network parameterized by $\theta$. The goal in RL is to find parameters which maximize the agent's ability to collect rewards. In a finite time-horizon, the goal is accomplished by finding CNN parameters $\theta^\star$:
\begin{equation}
\theta^{\star} = \argmax_\theta \sum_{t=0}^{T-1} r(s_t,a_t),
\label{eq:sum}
\end{equation}
where $T-1$ is the number of time steps experienced in the episode. In summary, the objective of our cube-collecting task is to find the CNN parameters to maximize the drone's ability to move toward cubes.

After each episode is finished, we use the \textbf{REINFORCE} algorithm to update the CNN parameters $\theta$ \cite{williams1992simple}. REINFORCE is an iterative algorithm which uses gradient decent to adjust $\theta$ in a direction which increases action probabilities that led to cube-collection in the prior episode. The amount of each adjustment is scaled by a \textbf{learning-rate}. This simple algorithm is ideal for the initial experiments presented here, because it allowed us to quickly understand the fundamental challenges in RL visualization.

We experiment with three algorithms to visualize the CNN policy's behavior: t-SNE, class visualization, attribution visualization.

\subsection{t-SNE}
\textbf{T-Distributed Stochastic Neighbor Embedding} (t-SNE) is a dimensionality reduction algorithm developed by \cite{maaten2008visualizing}. It is well suited for visualizing high-dimensional datasets. The method positions each high-dimensional datapoint (e.g. images) in a two or three-dimensional map in a way that similar datapoints are nearby and dissimilar ones are distant. The most recent use of t-SNE is to use a trained convolutional neural network (CNN) to extract features from each image, feed the features to t-SNE to get the position of each image, and arrange the images on a 2D or 3D space based on the given positions.

\subsection{Class Visualization}
Class visualization methods generate visual inputs which activate a particular output in a \textit{trained} neural network. This approach allows for a high-level of human comprehension about the behavior of a network, rather than treating the network like a black-box model. For our specific feature visualization approach, we use \textbf{Class Model Visualization} (CMV) \cite{simonyan2013deep}. CMV generates inputs which will trigger any specific output class in a trained convolutional neural network.

Formally, we let $a$ represent the action for which we want to generate an input image to trigger, $s$ is the input image which will be optimized such that the action probability is maximized. We let $\pi_\theta(a|s)$ represent the probability of taking action $a$ given the image $s$. The goal then is to solve the following optimization problem:
\begin{equation}
    s^* = \argmax_{s} \pi_\theta(a|s).
    \label{eq:cmv}
\end{equation}
In practice, the optimal image $s^*$ is found using gradient ascent by an automatic differentiation tool like TensorFlow.

\subsection{Attribution Visualization}

\textbf{Attribution visualization} techniques highlight regions in an input which are most responsible for a particular action in a CNN-based policy. We will use an attribution visualization technique called \textbf{Gradient-weighted Class Activation Mapping} (Grad-CAM) \cite{selvaraju2017grad}, which highlights regions in the input image most responsible for an action probability.

CNNs are particularly well-suited for attribution visualization, because they maintain spatial structure of the input as it flows through the network, this is why we can extract meaning from the last layer. A \textbf{feature map} is the output of a convolutional layer after it has passed through a nonlinearity (e.g. ReLU) function. Feature maps typically have many channels, and the goal of Grad-CAM is to find which channels contribute the most to an action taken. Grad-CAM achieves that goal by calculating the average derivative of the policy network, given a specific action $a$ and input image $s$, with respect to the feature map of interest:
\begin{equation}
    \alpha_k = \frac{1}{Z}\sum_i\sum_j \frac{\partial \pi_\theta(a|s)}{\partial A_{ij}^k},
    \label{eq:dfm}    
\end{equation}
where $A$ is the feature map of our target convolutional layer, $A^k$ is channel $k$ of $A$, $A_{ij}^k$ is the neuron at position $i,j$, and $Z=i \times j$. $\alpha_k$ is known as the \textbf{importance weight} for feature map channel $k$.

Once importance weights $\alpha_1,\alpha_2,\cdots,\alpha_K$ are known, they may be used to linearly combine feature map channels 1 through $K$, giving a ``class activation map'':
\begin{equation}
    \text{Grad-CAM} = \text{ReLU}\big(\sum_k{\alpha^k A^k}\big),
    \label{eq:gradcam}
\end{equation}
where ReLU is being used to filter for derivatives with a positive effect.

\begin{figure}
\centering
\subcaptionbox{High-performance policy}{%
  \includegraphics[width=0.45\textwidth]{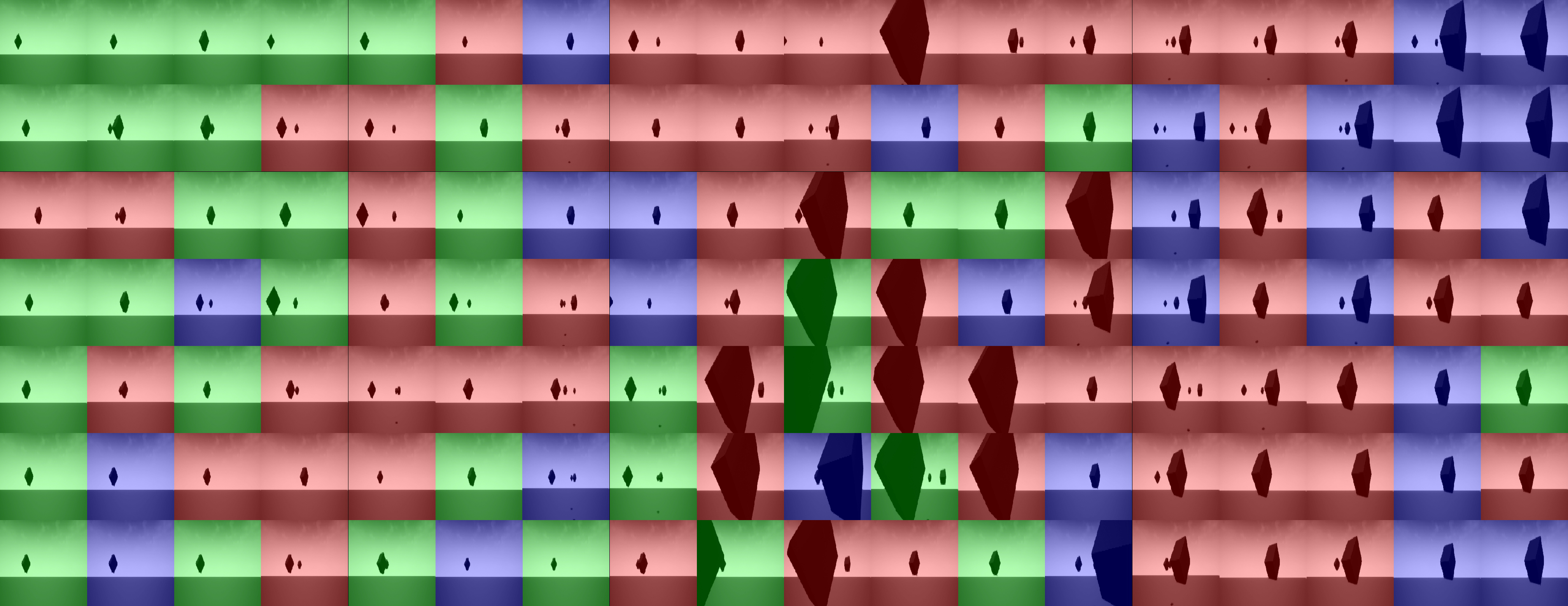}%
  }\par\medskip
\subcaptionbox{Poor-performance policy}{%
  \includegraphics[width=0.45\textwidth]{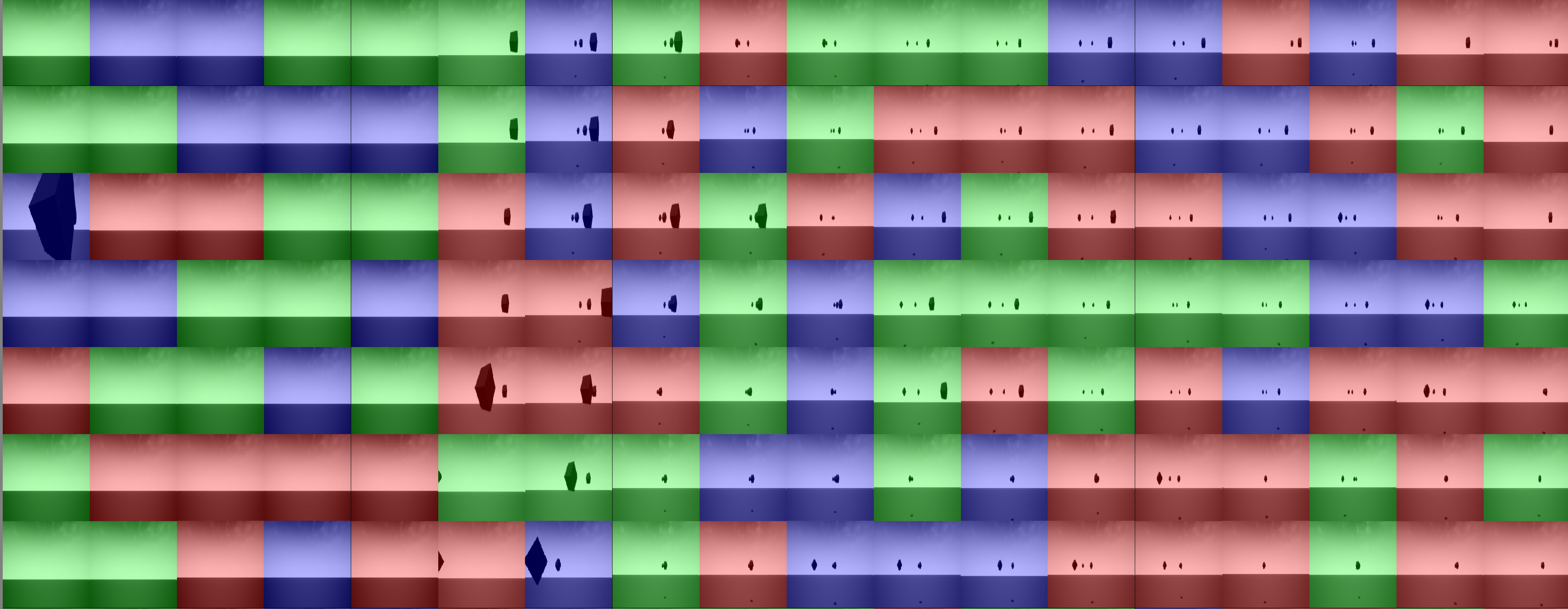}%
  }\par\medskip        
\subcaptionbox{Forward-and-right-only policy}{%
  \includegraphics[width=0.45\textwidth]{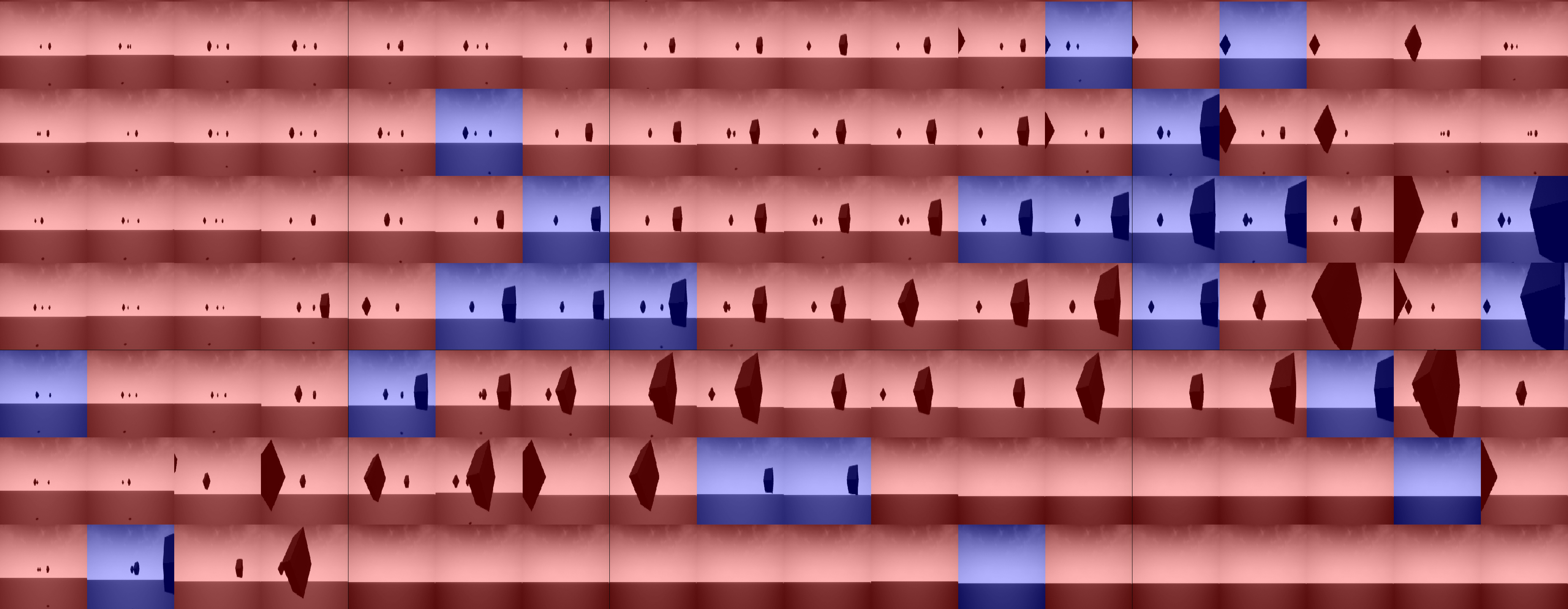}%
  }
\caption{T-SNE visualizations for the (a) high-performance policy, (b) poor-performance policy and (c) a policy which only moves right and forward. The tinting of each visual input is based on the action taken by the policy, given the input: red indicates ``forward'', green indicates ``left'', and blue indicates ``right''.}
\label{fig:tsne}
\end{figure}

\subsection{Related work}
There has only been a marginal amount of previous literature regarding the visualization of reinforcement learning policies. The creators of the DQN algorithm used t-SNE to cluster actions according to Atari game screens \cite{mnih2015human}. More in-depth exploration of t-SNE and Atari games was performed by \cite{zahavy2016graying}. Attribution visualization was used by \cite{greydanus2017visualizing} to analyze Atari video games. To the best of the authors' knowledge, this is the first work to apply t-SNE, class visualization, and attribution visualization to cyber-physical systems.

\section{Results}
We trained three separate policies for the cube-collection task. A \textbf{high-performance} policy was trained such that it almost always collected all cubes. A \textbf{poor-performance} policy was trained only briefly so it never learned. And a ``broken'' \textbf{forward-and-right-only} policy was trained to collect only boxes in front of it or to the right, and avoiding any box to the left. We applied t-SNE, class visualization, and attribution visualization to each of the policies to demonstrate how these tools aid in policy development and understanding.

\subsection{t-SNE}

Fig.~\ref{fig:tsne} provides t-SNE visualizations of the three policies. The tinting of each patch is based on the action taken by the policy, given the drone's observation: red indicates ``forward'', green indicates ``left'', and blue indicates ``right''. 

In Fig.~\ref{fig:tsne}(a) we observe from the high-performance policy that the visual inputs with the same tinted color are generally clustered together by the content of the image.\footnote{Clustering is not guaranteed because the policy is stochastic.} In comparison, the poor-performance policy Fig.~\ref{fig:tsne}(b) scatters colors all over the map. This is because each action is equally likely, and uncorrelated with the drone's observation. In Fig.~\ref{fig:tsne}(c) the forward-and-right-only policy is insightful, as there are no green (``left'') patches. It can also be observed that Fig.~\ref{fig:tsne}(c) shows mostly ``forward'' (red) actions and fewer ``right'' (blue) actions.
\subsection{Class Visualization}

\begin{figure}[t]
\subcaptionbox{High-performance policy}{%
\includegraphics[width = 1.125in]{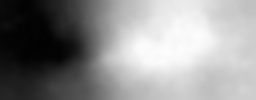}%
\includegraphics[width = 1.125in]{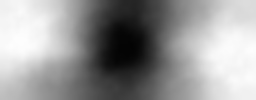}%
\includegraphics[width = 1.125in]{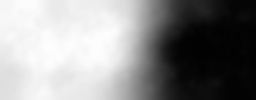}%
}\par\medskip  
\subcaptionbox{Poor-performance policy}{%
\includegraphics[width = 1.125in]{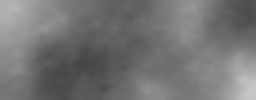}%
\includegraphics[width = 1.125in]{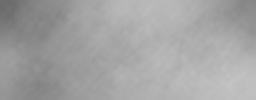}%
\includegraphics[width = 1.125in]{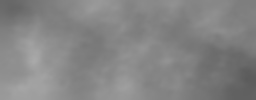}%
}\par\medskip  
\subcaptionbox{Forward-and-right-only policy}{%
\includegraphics[width = 1.125in]{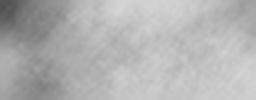}%
\includegraphics[width = 1.125in]{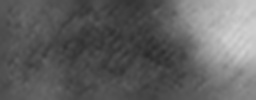}%
\includegraphics[width = 1.125in]{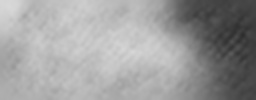}%
}
\caption{Class visualizations for the (a) high-performance policy, (b) poor-performance policy and (c) a policy which only moves right and forward. Generated images maximize action probabilities for their respective policies. The left-most image triggers the ``left'' drone action, middle triggers ``forward'', and right triggers ``right''.}
\label{fig:classviz}
\end{figure}

The high-performance policy's class visualization in Fig.~\ref{fig:classviz}(a) clearly explains what the policy is looking for, where the bias towards the ``left'', ``forward'', or ``right'' depends on the position of the cube. Specifically, if a box blocks the view of the camera on the left, then the policy will most likely choose the ``left'' action. Similarly for the ``forward'' and ``right'' actions.

Class visualization of the poor-performance policy is also insightful. Fig.~\ref{fig:classviz}(b) illustrates that the actions of the poorly trained model are equally triggered by noise.

Fig.~\ref{fig:classviz}(c) visualizes the forward-and-right-only policy, where only the ``right'' action visualization makes sense. That is, when the camera is occluded on the right, the policy will choose to move to the right. The ``left'' action shows that there is a small response to camera occlusion on the left, but the ``forward'' action dominates the left action. 

Class visualizations for the forward-and-right-only policy highlight one of the challenges in reinforcement learning. If the learning-rate in REINFORCE is too high, the policy might finalize its decision making process based on early experience. In this case, the drone experienced an early success by moving ``right'' and ``forward'', which resulted in the elimination of ``left'' action probabilities. The remedy for this was to lower the learning-rate of the policy updates.

\subsection{Attribution Visualization}

\begin{figure}[t]
\begin{tabular}{@{}c@{}c@{}}
\includegraphics[width = 1.7in]{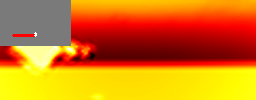}&
\includegraphics[width = 1.7in]{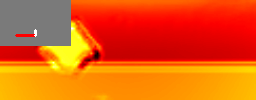}\\[-3pt]
\includegraphics[width = 1.7in]{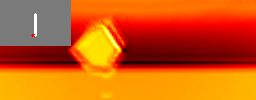}&
\includegraphics[width = 1.7in]{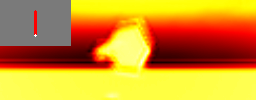}\\[3.75pt]
\multicolumn{2}{c}{(a) High-performance policy}\\[3.75pt]
%1&
%2\\[4pt]
\includegraphics[width = 1.7in]{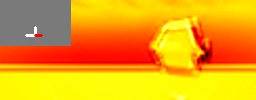}&
\includegraphics[width = 1.7in]{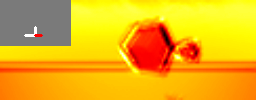}\\[-3pt]
\includegraphics[width = 1.7in]{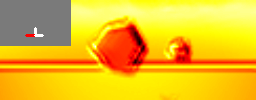}&
\includegraphics[width = 1.7in]{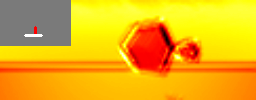}\\[3.75pt]
\multicolumn{2}{c}{(b) Poor-performance policy}\\[3.75pt]
\includegraphics[width = 1.7in]{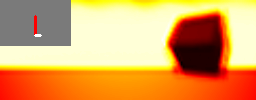}&
\includegraphics[width = 1.7in]{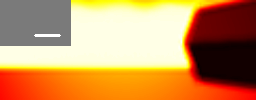}\\[-3pt]
\includegraphics[width = 1.7in]{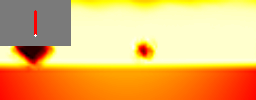}&
\includegraphics[width = 1.7in]{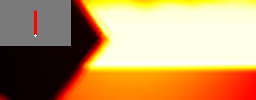}\\[3.5pt]
\multicolumn{2}{c}{(c) Forward-and-right-only policy}\\
\end{tabular}
\caption{Attribution visualizations for state observations and actions taken with the (a) high-performance policy, (b) poor-performance policy, and (c) a policy which only moves right and forward. The T-shape in the corner image visualizes action probabilities, and the red color indicates action taken.}
\label{fig:attvizr}
\end{figure}

Fig.~\ref{fig:attvizr} provides attribution visualizations. The T-shape in the corner of each image visualizes action probabilities, and the red color indicates the (stochastic) action taken. The bright areas of each image indicate image features which most contributed to the probability of the action taken.

Observing the upper-left image in Fig.~\ref{fig:attvizr}(a), we see that the ``right'' action was the only action with a high probability. Furthermore, the closest cube is the brightest (indicating its importance for the decision), followed by the second cube. Also note that the lower-left image visualizes what contributed to the low-probability ``left'' action which was made. 

In Fig.~\ref{fig:attvizr}(b) we see visualizations for the poor-performance policy. In that figure, observe that all action probabilities are roughly the same, as indicated by the T-shape, regardless of the position of the drone relative to the cubes. 

Fig.~\ref{fig:attvizr}(c) provides attribution visualizations for a model that will only move ``right'' and ``forward''. Notice that the colors here are inverted compared to Fig.~\ref{fig:attvizr}(a), which is odd. The policy pays more attention to the horizon than anything else. Observe that the lower-left image shows no ``left'' reaction to the cube on the left.

\section{Conclusions and future work}

We see an increasing number of businesses using convolutional neural networks in vision-based cyber-physical systems. Simultaneously, the research community has been actively coupling CNNs with traditional reinforcement learning algorithms. We may soon begin to interact with RL-enabled CPS in our day-to-day lives. But this is currently dangerous because of the opacity of CNNs. 

Our research is a step in the direction of creating understandable and trustworthy vision-based RL systems through policy visualization. We have adapted three existing CNN visualization techniques to our drone simulation environment: t-SNE, class visualization, and attribution visualization. But the adaptation of existing CNN visualization techniques addressed here are not adequate on their own.

Numerous opportunities exist for advancing this domain. For example reinforcement learning is inherently time-series based, it often makes stochastic decisions, and real-time visualization would be useful for some applications. Our future efforts will explore these avenues.

% or


%\appendices

%\section*{Acknowledgment}


\ifCLASSOPTIONcaptionsoff
  \newpage
\fi



%\bibliographystyle{IEEEtran}
%\bibliographystyle{IEEEabrv}
%\bibliographystyle{unsrt}
%\bibliography{references}
\bibliographystyle{IEEEtran}
\bibliography{references}

%\bibliography{IEEEexample}


%\vfill


\end{document}